\documentclass[runningheads]{llncs}
\usepackage[dvipdfmx]{graphicx}
\usepackage{graphicx}
\usepackage{amsmath,amssymb} 
\usepackage{color}
\usepackage{bm}
\usepackage[width=122mm,left=12mm,paperwidth=146mm,height=193mm,top=12mm,paperheight=217mm]{geometry}

\newcommand{\Tref}[1]{Table~\ref{#1}}
\newcommand{\Eref}[1]{Eq.~(\ref{#1})}
\newcommand{\Fref}[1]{Fig.~\ref{#1}}
\newcommand{\Sref}[1]{Section~\ref{#1}}
\newcommand{\Cref}[1]{Chapter~\ref{#1}}
\newcommand{\eg}[0]{\emph{e.g}.}
\newcommand{\ie}[0]{\emph{i.e}.}
\newcommand{\etal}[0]{\emph{et al}.}

\begin{document}
\pagestyle{headings}
\mainmatter

\title{CNN-PS: CNN-based Photometric Stereo for General Non-Convex Surfaces} 

\titlerunning{CNN-PS}

\authorrunning{S. Ikehata}

\author{Satoshi Ikehata}


\institute{	National Institute of Informatics, Tokyo, Japan\\
	\email{sikehata@nii.ac.jp}
}
\maketitle

\begin{abstract}
Most conventional photometric stereo algorithms inversely solve a BRDF-based image formation model. However, the actual imaging process is often far more complex due to the global light transport on the non-convex surfaces. This paper presents a photometric stereo network that directly learns relationships between the photometric stereo input and surface normals of a scene. For handling unordered, arbitrary number of input images, we merge all the input data to the intermediate representation called {\it observation map} that has a fixed shape, is able to be fed into a CNN. To improve both training and prediction, we take into account the rotational pseudo-invariance of the observation map that is derived from the isotropic constraint. For training the network, we create a synthetic photometric stereo dataset that is generated by a physics-based renderer, therefore the global light transport is considered. Our experimental results on both synthetic and real datasets show that our method outperforms conventional BRDF-based photometric stereo algorithms especially when scenes are highly non-convex.
\end{abstract}
\section{Introduction} \label{sec:introduction}
In 3-D computer vision problems, the input data is often {\it unstructured} (\ie, the number of input images is varying and the images are unordered). A good example is the multi-view stereo problem where the scene geometry is recovered from unstructured multi-view images. Due to this unstructuredness, 3-D reconstruction from multiple images less relied on the supervised learning-based algorithms except for some structured problems such as binocular stereopsis~\cite{Kendall2017} and two-view SfM~\cite{Vijayanarasimhan2017} whose number of input images is always fixed. However, recent advances in deep convolutional neural network (CNN) have motivated researchers to address unstructured 3-D computer vision problems with deep neural networks. For instance, a recent work from Kar~\etal~\cite{Kar2017} presented an end-to-end learned system for the multi-view stereopsis while Kim~\etal~\cite{Kim2017} presented a learning-based surface reflectance estimation from multiple RGB-D images. Either work intelligently merged all the unstructured input to a structured, intermediate representation (\ie, 3-D feature grid~\cite{Kar2017} and 2-D hemispherical image~\cite{Kim2017}). 
\renewcommand{\thefootnote}{}
\footnote{This work was supported by JSPS KAKENHI Grant Number JP17H07324.}
\renewcommand{\thefootnote}{\arabic{footnote}}

Photometric stereo is another 3-D computer vision problem whose input is unstructured, where surface normals of a scene are recovered from appearance variations under different illuminations. Photometric stereo algorithms typically solved an inverse problem of the pointwise image formation model which was based on the Bidirectional Reflectance Distribution Function (BRDF). While effective, a BRDF-based image formation model generally cannot account the global illumination effects such as shadows and inter-reflections, which are often problematic to recover non-convex surfaces. Some algorithms attempted the robust outlier rejection to suppress the non-Lambertian effects~\cite{Wu2010,ikehata2012,ikehata2014a,Queau2017}, however the estimation failed when the non-Lambertian observation was dominant. This limitation inevitably occurs due to the fact that multiple interactions of light and a surface are difficult to be modeled in a mathematically tractable form.

To tackle this issue, this paper presents an end-to-end CNN-based photometric stereo algorithm that learns the relationships between surface normals and their appearances without physically modeling the image formation process. For better scalability, our approach is still pixelwise and rather inherit from conventional robust approaches~\cite{Wu2010,ikehata2012,ikehata2014a,Queau2017}, which means that we learn the network that automatically ``neglects" the global illumination effects and estimate the surface normal from ``inliers" in the observation. To achieve this goal, we will train our network on as much as possible synthetic patterns of the input that is ``corrupted" by global effects. Images are rendered with different complex objects under the diverse material and illumination condition. 

Our challenge is to apply the deep neural network  to the photometric stereo problem whose input is unstructured. In similar with recent works~\cite{Kar2017,Kim2017}, we merge all the photometric stereo data to an intermediate representation called {\it observation map} that has a fixed shape, therefore is naturally fed to a standard CNN. As many photometric stereo algorithms were, our work is also primarily concerned with isotropic materials, whose reflections are invariant under rotation about the surface normal. We will show that this isotropy can be taken advantages of in a form of the {\it rotational pseudo-invariance}  of the observation map for both augmenting the input data and reducing the prediction errors. To train the network, we create a synthetic photometric stereo dataset ({\it CyclesPS}) by leveraging the physics-based Cycles renderer~\cite{Cycles} to simulate the complex global light transport. For covering diverse real-world materials, we adopt the Disney's principled BSDF~\cite{DisneyPrincipledBSDF} that was proposed for artists to render various scenes by controlling small number of parameters.

We evaluate our algorithm on the DiLiGenT Photometric Stereo Dataset~\cite{Shi2018} which is a real benchmark dataset containing images and calibrated lightings. We compare our method against conventional photometric stereo algorithms~\cite{Woodham1980,Alldrin2008,Goldman2010,Higo2010,Wu2010,ikehata2012,Shi2012b,ikehata2014a,ikehata2014b,Shi2014,Queau2017,Santo2017,Hui2017,taniai2018} and show that our end-to-end learning-based algorithm most successfully recovers the non-convex, non-Lambertian surfaces among all the algorithms concerned. 

The summary of contributions is following:\\
\noindent(1) We firstly propose a supervised CNN-based calibrated photometric stereo algorithm that takes unstructured images and lighting information as input. \\
\noindent(2) We present a synthetic photometric stereo dataset ({\it CyclesPS}) with a careful injection of the global illumination effects such as cast shadows, inter-reflections.\\
\noindent(3) Our extensive evaluation shows that our method performs best on the DiLiGenT benchmark dataset~\cite{Shi2018} among various conventional algorithms especially when the surfaces are highly non-convex and non-Lambertian.

Henceforth we rely on the classical assumptions on the photometric stereo problem (\ie, fixed, linear orthographic camera and known directional lighting). 

\section{Related Work}
Diverse appearances of real world objects can be encoded by a BRDF $\rho$, which relates the observed intensity $I_{j}$ to the associated surface normal $\bm{n} \in \mathbb{R}^3$, the $j$-th incoming lighting direction $\bm{l}_j \in \mathbb{R}^3$, its intensity $L_j\in\mathbb{R}$, and the outgoing viewing direction $\bm{v} \in \mathbb{R}^3$ via
\begin{equation}
I_{j} = L_j\rho(\bm{n},\bm{l}_j,\bm{v})\max{(\bm{n}^{\top}\bm{l}_j,0)} + \epsilon_{j}, \label{eq:img_form1}
\end{equation}
where $\max{(\bm{n}^{\top}\bm{l}_j,0)}$ accounts for attached shadows and $\epsilon_j$ is an additive error to the model. \Eref{eq:img_form1} is generally called {\it image formation model}. Most photometric stereo algorithms assumed the specific shape of $\rho$ and recovered the surface normals of a scene by inversely solving~\Eref{eq:img_form1} from a collection of observations under $m$ different lighting conditions $(j\in 1,\cdots,m)$. All the effects that are not represented by a BRDF (image noises, cast shadows, inter-reflections and so on) are typically put together in $\epsilon_j$. Note that when the BRDF is Lambertian and the additive error is removed, it is simplified to the traditional Lambertian image formation model~\cite{Woodham1980}.

Since Woodham firstly introduced the Lambertian photometric stereo algorithm, the extension of its work to non-Lambertian scenes has been a problems of significant interest. Photometric stereo approaches to dealing with non-Lambertian effects are mainly categorized into four classes: (a) robust approach, (b) reflectance modeling with non-Lambertian BRDF, (c) example-based reflectance modeling and (d) learning-based approach. 

Many photometric stereo algorithms recover surface normals of a scene via a simple diffuse reflectance modeling (\eg, Lambertian) while treating other effects as outliers. For instance, Wu~\etal~\cite{Wu2010} have proposed a rank-minimization based approach to decompose images into the low-rank Lambertian image and non-Lambertian sparse corruptions. Ikehata~\etal~ extended their method by constraining the rank-3 Lambertian structure~\cite{ikehata2012} (or the general diffuse structure~\cite{ikehata2014a}) for better computational stability. Recently, Queau~\etal~\cite{Queau2017} have presented a robust variational approach for inaccurate lighting as well as various non-Lambertian corruptions. While effective, a drawback of this approach is that if it were not for dense diffuse inliers, the estimation fails. 

Despite their computational complexity, various algorithms arrange the parametric or non-parametric models of non-Lambertian BRDF. In recent years, there has been an emphasis on representing a material with a small number of fundamental BRDF. Goldman~\etal~\cite{Goldman2005} have approximated each fundamental BRDF by the Ward model~\cite{Ward1992} and Alldrin~\etal~\cite{Alldrin2008} later extended it to non-parametric representation. Since the high-dimensional ill-posed problem may cause the instability of the estimation, Shi~\etal~\cite{Shi2014} presented a compact biquadratic representation of isotropic BRDF. On the other hand, Ikehata~\etal~\cite{ikehata2014b} introduced the sum-of-lobes isotropic reflectance model~\cite{Chandraker2011a} to account all frequencies in isotropic observations. For improving the efficiency of the optimization, Shen~\etal~\cite{Shen2017} presented a kernel regression approach, which can be transformed to an eigen decomposition problem. This approach works well as far as a resultant image formation model is correct without model outliers.

A few amount of photometric stereo algorithms are grouped into the {\it example-based} approach, which takes advantages of the surface reflectance of objects with known shape, captured under the same illumination environment with the target scene. The earliest example-based approach~\cite{Silver1980} requires a reference object whose material is exactly same with that of target object. Hertzmann~\etal~\cite{Hertzmann2005} have eased this restriction to handle uncalibrated scenes and spatially varying materials by assuming that materials 
can be expressed as a small number of basis materials. Recently, Hui~\etal~\cite{Hui2017} presented an example-based method without a physical reference object by taking advantages of virtual spheres rendered with various materials. While effective, this approach also suffers from model outliers and has a drawback that the lighting configuration of the reference scene must be taken over at the target scene. 

Machine learning techniques have been applied in a few very recent photometric stereo works~\cite{taniai2018,Santo2017}. Santo~\etal~\cite{Santo2017} presented a supervised learning-based photometric stereo method using a neural network that takes as input a normalized vector where each element corresponds to an observation under specific illumination. A surface normal is predicted by feeding the vector to one dropout layer and adjacent six dense layers. While effective, this method has limitation that lightings remain the same between training and test phases, making it inapplicable to the unstructured input. One another work by Taniai and Maehara~\cite{taniai2018} presented an unsupervised learning framework where surface normals and BRDFs are predicted by the network trained by minimizing reconstruction loss between observed and synthesized images with a rendering equation. While their network is invariant to the number and permutation of the images, the rendering equation is still based on a point-wise BRDF and intolerant to the model outliers. Furthermore, they reported slow running time (\ie, 1 hour to do 1000 SGD iterations for each scene) due to its self-supervision manner.

In summary, there is still a constant struggle in the design of the photometric stereo algorithm among its complexity, efficiency, stability and robustness. Our goal is to solve this dilemma. Our end-to-end learning-based algorithm builds upon the deep CNN trained on synthetic datasets, abandoning the modeling of complicated image formation process. Our network accepts the unstructured input (\ie, our network is invariant to both number and order of input images) and works for various real-world scenes where non-Lambertian reflections are intermingled with global illumination effects. 
\section{Proposed Method}
Our goal is to recover surface normals of a scene of (a) spatially-varying isotropic materials and with (b) global illumination effects (\eg, shadows and interreflections) (c) where the scene is illuminated by unknown number of lights. To achieve this goal, we propose a CNN architecture for the calibrated photometric stereo problem which is invariant to both the number and order of input images. The tolerance to global illumination effects is learned from the synthetic images of non-convex scenes rendered with the physics-based renderer.
\subsection{2-D observation map for unstructured photometric stereo input}
\begin{figure}[!t]
	\begin{center}
		\includegraphics[width=120mm]{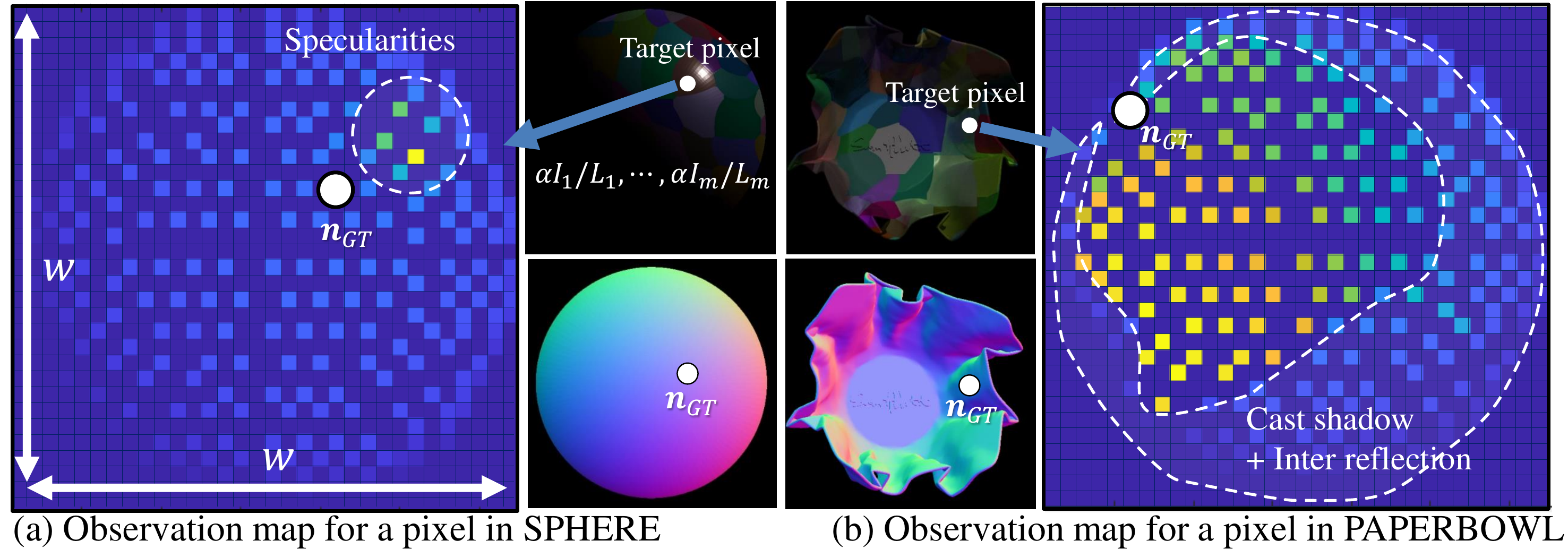}
	\end{center}
	\vspace{-5mm}
	\caption{We project pairs of images and lightings to a fixed-size observation map based on the bijective mapping of a light direction from a hemisphere to the 2-D coordinate system perpendicular to the viewing axis. This figure shows observation maps for (a) a point on a smooth convex surface and (b) a point on a rough non-convex surface. We also projected the true surface normal at the point onto the same coordinate system of the observation map for reference.}
	\label{fig:2dmap}
\end{figure}
We firstly present the {\it observation map} which is generated by a pixelwise hemispherical projection of observations based on known lighting directions. Since a lighting direction is a vector spanned on a unit hemisphere, there is a bijective mapping from $\bm{l}_j\triangleq[l^j_{x}\;l^j_{y}\;l^j_{z}]^{\top}\in \mathbb{R}^3$  to $[l^j_{x}\;l^j_{y}]^{\top}\in \mathbb{R}^2$ (s.t., $l^2_x + l^2_y + l^2_z = 1$) by projecting a vector onto the $x$-$y$ coordinate system which is perpendicular to a viewing direction ($\ie, \bm{v} = [0\;0\;1]$).\footnote{We preliminarily tried the projection on the spherical coordinate system ($\theta,\phi$), but the performance was worse than one on the standard x-y coordinate system.} Then we define an observation map $O\in \mathbb{R}^{w\times w}$ as
\begin{equation}
O_{{\rm int}(w(l_{x}+1)/2), {\rm int}(w(l_{y}+1)/2)} = \alpha I_j/L_j\;\;\forall \;j\in\;1,\cdots,m,\label{eq:projection}
\end{equation}
where ``int" is an operator to round a floating value to an integer and $\alpha$ is a scaling factor to normalize data (\ie, we simply use $\alpha={\rm max}\;L_j/I_j$). Once all the observations and lightings are stored in the observation map, we take it as an input of the CNN. Despite its simplicity, this representation has three major benefits. First, its shape is independent of the number and size of input images. Second, the projection of observations is order-independent (\ie, the observation map does not change when swapping $i$-th and $j$-th images). Third, it is unnecessary to explicitly feed the lighting information into the network.

\Fref{fig:2dmap} illustrates examples of the observation map of two objects namely SPHERE and PAPERBOWL, one is purely convex and the other is highly non-convex. \Fref{fig:2dmap}-(a) indicates that the target point could be on the convex surface since the values of the observation map gradually decrease to zero as the light direction is going apart from the true surface normal ($\bm{n}_{GT}$). The local concentration of large intensity values also indicates the narrow specularity on the smooth surface. On the other hand, the abrupt change of values in \Fref{fig:2dmap}-(b) evidences the presence of cast shadows or inter-reflections on the non-convex surface. Since there is no local concentration of intensity values, the surface is likely to be rough. In this way, an observation map reasonably encodes the geometry, material and behavior of the light at around a surface point.

\begin{figure}[!t]
	\begin{center}
		\includegraphics[width=120mm]{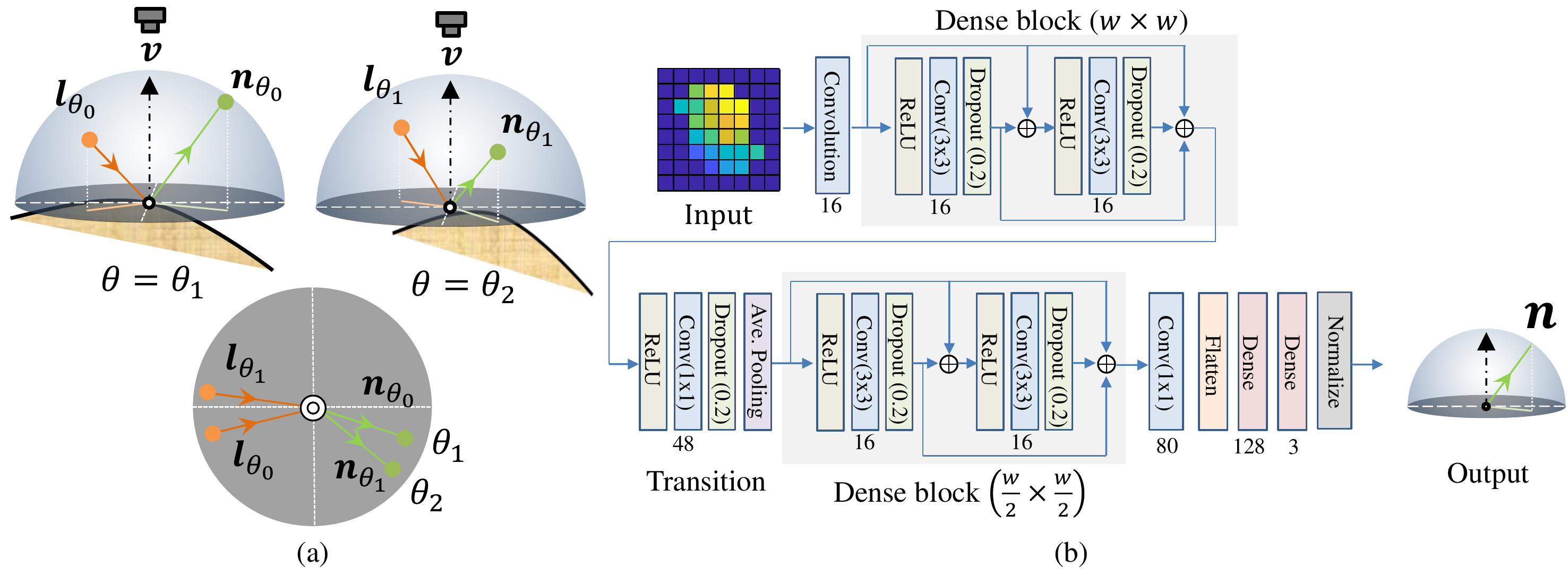}
	\end{center}
	\vspace{-5mm}
	\caption{(a) Isotropy guarantees that the appearance of a surface from $\bm{v}$ is invariant of the rotation of $\bm{l}$ and $\bm{n}$ around the view axis. (b) Our network architecture is a variation of DenseNet~\cite{Huan2017} that outputs a normalized surface normal from a $32\times32$ observation map. Numbers of the filter are presented below each layer.}
	\label{fig:hemi}
\end{figure}

\subsection{Rotation pseudo-invariance for the isotropy constraint}
An observation map $O$ is sparse in a general photometric stereo setup (\eg, assuming that $w=32$ and we have 100 images as input, the ratio of non-zero entries in $O$ is about $10\%$). The missing data is generally considered problematic as CNN input and often interpolated~\cite{Kim2017}. However, we empirically found that smoothly interpolating missing entries degrades the performance since an observation map is often non-smooth and zero values have an important meaning (\ie, shadows). Therefore we alternatively try to improve the performance by taking into account the isotropy of the material. 

Many real-world materials exhibit identically same appearance when the surface is rotated along a surface normal. The presence of this behavior is referred to as isotropy~\cite{Matusik2003,Alldrin2007b}. Isotropic BRDFs are parameterized in terms of three values instead of four~\cite{Stark2005} as  
\begin{equation}
\rho = f(\bm{n}^{\top}\bm{l},\bm{n}^{\top}\bm{v},\bm{l}^{\top}\bm{v}), \label{eq:isotropy}
\end{equation}
where $f$ is an arbitrary reflectance function.\footnote{Note that there are other parameterizations of an isotropic BRDF~\cite{Montes2012}.} Combining~\Eref{eq:isotropy} with~\Eref{eq:img_form1}, we get following image formation model.
\begin{equation}
I = Lf(\bm{n}^{\top}\bm{l},\bm{n}^{\top}\bm{v},\bm{l}^{\top}\bm{v})\max{(\bm{n}^{\top}\bm{l},0)}. \label{eq:img_form2}
\end{equation}
Note that lighting index and model error are omitted for brevity. Let's consider the rotation of surface normal $\bm{n}$ and lighting direction $\bm{l}$ around the z-axis (\ie, viewing axis) as $\bm{n}^{\prime} = [(R[n_x\;n_y]^{\top})^{\top}\;n_z]^{\top}, \bm{l}^{\prime} = [(R[l_x\;l_y]^{\top})^{\top}\;l_z]^{\top}$ where $\bm{n}\triangleq [n_x\;n_y\;n_z]^{\top}$ and $R \in SO(2)$ is an arbitrary rotation matrix. Then,
\begin{eqnarray}
{\bm{n}^{\prime}}^{\top}\bm{l}^{\prime} &=&[(R[n_x\;n_y]^{\top})^{\top}\;n_z][(R[l_x\;l_y]^{\top})^{\top}\;l_z]^{\top}\\ \nonumber
&=&[n_x\;n_y]R^{\top}R[l_x\;l_y]^{\top} + n_zl_z=\bm{n}^{\top}\bm{l},\\
{\bm{n}^{\prime}}^{\top}\bm{v}^{\prime} &=&[(R[n_x\;n_y]^{\top})^{\top}\;n_z][0\;0\;1]^{\top}=n_z = \bm{n}^{\top}\bm{v},\\
{\bm{l}^{\prime}}^{\top}\bm{v}^{\prime} &=&[(R[l_x\;l_y]^{\top})^{\top}\;l_z][0\;0\;1]^{\top}=l_z = \bm{l}^{\top}\bm{v}.
\end{eqnarray}
Feeding them into~\Eref{eq:img_form2} gives following equation,
\begin{eqnarray}
I&=&Lf({\bm{n}^{\prime}}^{\top}\bm{l^{\prime}},{\bm{n^{\prime}}}^{\top}\bm{v},{\bm{l^{\prime}}}^{\top}\bm{v})\max{({\bm{n^{\prime}}}^{\top}\bm{l^{\prime},0)}}\label{eq:invariance}\\\nonumber
&=&Lf(\bm{n}^{\top}\bm{l},\bm{n}^{\top}\bm{v},\bm{l}^{\top}\bm{v})\max{(\bm{n}^{\top}\bm{l},0)}.
\end{eqnarray}
Therefore, the rotation of lighting and surface normal around $z$-axis does not change the appearance as illustrated in~\Fref{fig:hemi}-(a). Note that this theorem holds even for the indirect illumination in non-convex scenes by rotating all the geometry and environment illumination around the viewing axis. This result is important for our CNN-based algorithm. We suppose that a neural network is a mapping function $g: x \mapsto g(x)$ that maps $x$ (\ie, a set of images and lightings) to $g(x)$ (\ie, a surface normal) and $r$ is a rotation operator of lighting/normal at the same angle around $z$-axis. From~\Eref{eq:invariance}, we get $r(g(x))=g(r(x))$. We call this relationship as {\it rotational pseudo-invariance} (the standard rotation invariance is $g(x)=g(r(x))$). Note that this rotational pseudo-invariance is also applied on the observation map since the rotation of lightings around the viewing axis results in the rotation of the observation map around the z-axis\footnote{Strictly speaking, we rotate the lighting directions instead of the observation map itself. Therefore, we do not need to suffer from the boundary issue unlike the standard rotational data augmentation.}.

We constrain the network with the rotational pseudo-invariance in the similar manner that the rotation invariance is achieved. Within the CNN framework, two approaches are generally adopted to encode the rotation invariance. One is applying rotations to the input image~\cite{Simard2003} and the other is applying rotations to the convolution kernels~\cite{Schmidt2012}. We adopt the first strategy due to its simplicity. Concretely, we augment the training set with many rotated versions of lightings and surface normal, which allows the network to learn the invariance without explicitly enforcing it. In our implementation, we rotate the vectors at $10$ regular intervals from 0 to 360.
\begin{figure}[!t]
	\begin{center}
		\includegraphics[width=120mm]{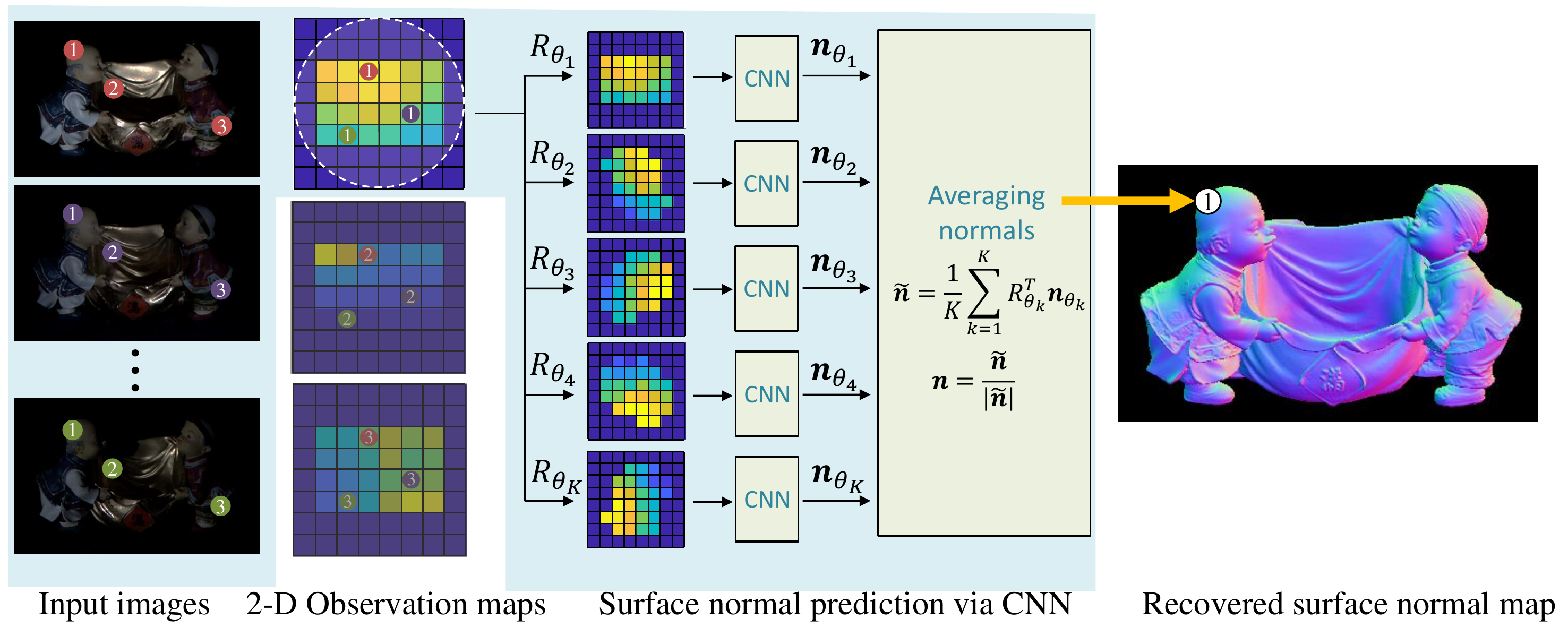}
	\end{center}
	\vspace{-5mm}
	\caption{The illustration of the prediction module. For each surface point, we generate $K$ observation maps taking into account the rotational pseudo-invariance. Each observation map is fed into the network and all the output normals are averaged.}
	\label{fig:prediction}
\end{figure}
\subsection{Architecture details} 
In this section, we describe the framework of training and prediction. Given images and lightings, we produce observation maps followed by~\Eref{eq:projection}. Data is augmented to achieve the rotational pseudo-invariance by rotating both lighting and surface normal vectors around the viewing axis. Note that a color image is converted to a gray-scale image. The size of the observation map ($w$) should be chosen carefully. As $w$ increases, the observation map becomes sparser. On the other hand, the smaller observation map has less respresentability. Considering this trade-off, we empirically found that $w=32$ is a reasonable choice (we tried $w=8,16,32,64$ and $w=32$ showed the best performance when the number of images is less than one thousand). 

A variation of densely connected convolutional neural network (DenseNet~\cite{Huan2017}) architecture is used to estimate a surface normal from an observation map. The network architecture is shown in~\Fref{fig:hemi}-(b). The network includes two 2-layer dense blocks, each consists of one activation layer (relu), one convolution layer ($3\times3$) and a dropout layer ($20 \%$ drop) with a concatenation from the previous layers. Between two dense blocks, there is a transition layer to change feature-map sizes via convolution and pooling. We do not insert a batch normalization layer that was found to degrade the performance in our experiments. After the dense blocks, the network has two dense layers followed by one normalization layer which convert a feature to an unit vector. The network is trained with a simple mean squared loss between predicted and ground truth surface normals. The loss function is minimized using Adam solver~\cite{Adam}. We should note that since our input data size is relatively small (\ie, $32\times 32 \times 1$), the choice of the network architecture is not a critical component in our framework.\footnote{We compared architectures of AlexNet, VGG-NET and densenet as well as much simpler architectures with only two or three convolutoinal layers and the dense layer(s). Among the architectures we tested, the current architecture was slightly better.} 

The prediction module is illustrated in~\Fref{fig:prediction}. Given observation maps, we predict surface normals based on the trained network. Since it is practically impossible to train the perfect rotational pseudo-invariant network, estimated surface normals for differently rotated observation maps were not identical (typically the difference of angular errors between every two different rotations was less than 10\%-20\% of their average). For further emphasizing the rotational pseudo-invariance, we again augment the input data by rotating lighting vectors at a certain angle $\theta \in \theta_1,\cdots \theta_K$ and then merge the outputs into one. Suppose the surface normal ($\bm{n}_{\theta}$) is a prediction from the input data rotated by $R_{\theta}$, then we simply average the inversely rotated surface normals as follows,
\begin{eqnarray}
\bar{\bm{n}} &=& \frac{1}{K}\sum_{k=1}^K{R^{\top}_{\theta_k}\bm{n}_{\theta_k}},\label{eq:prediction}\\
\bm{n} &=& \bar{\bm{n}}/\|\bar{\bm{n}}\|.\nonumber
\end{eqnarray}
\subsection{Training dataset ({\it CyclesPS} dataset)}
In this section, we present our {\it CyclesPS} training dataset. {\it DiLiGenT}~\cite{Shi2018}, the largest real photometric stereo dataset contains only ten scenes with fixed lighting configuration. Some works~\cite{Shi2014,ikehata2014b,Santo2017} attempted to synthesize images with MERL BRDF database~\cite{Matusik2003}, however only one hundred measured BRDFs cannot cover the tremendous real-world materials. Therefore, we decided to create our own training dataset that has diverse materials, geometries and illumination. 

For rendering scenes, we collected high quality 3-D models under royalty free license  from the internet.\footnote{References to each 3-D model are included in supplementary.} We carefully chose fifteen models for training and three models for test whose surface geometry is sufficiently complex to cover the diverse surface normal distribution. Note that we empirically found 3-D models in ShapeNet~\cite{shapenet2015} which was used in a previous work~\cite{Kim2017} are generally too simple (\eg, models are often low-polygonal, mostly planar) to train the network.

\begin{figure}[!t]
	\begin{center}
		\includegraphics[width=120mm]{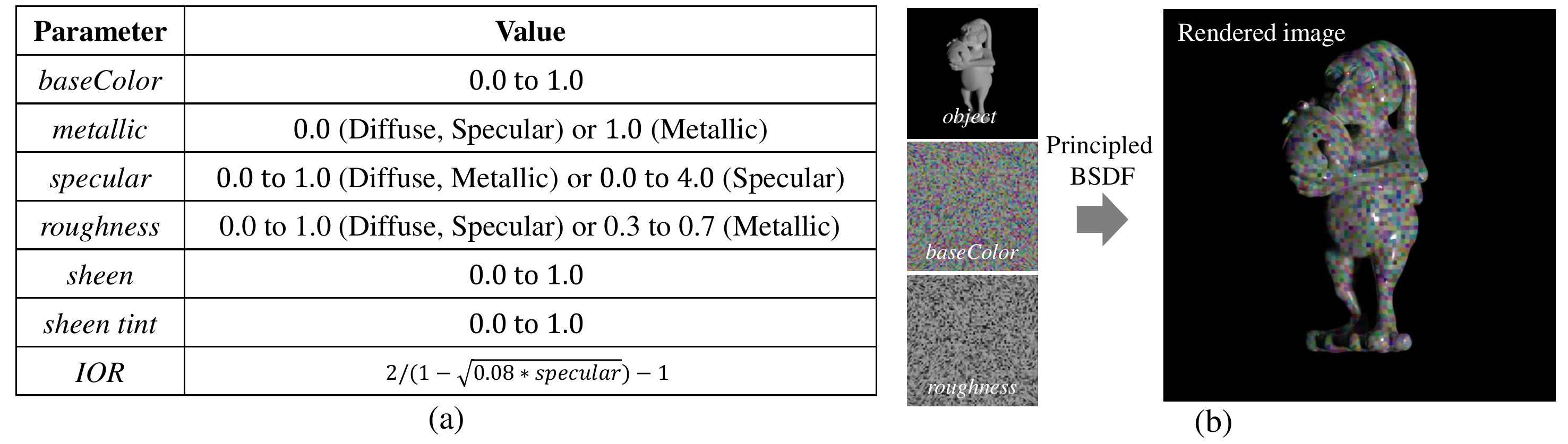}
	\end{center}
	\vspace{-5mm}
	\caption{ (a) The range of each parameter in the principled BSDF~\cite{DisneyPrincipledBSDF} is restricted by three different material configurations (Diffuse, Specular, Metallic). (b) The material parameters are passed to the renderer in the form of a 2-D texture map.}
	\label{fig:datageneration}
\end{figure}
The representation of the reflectance is also important to make the network robust to wide varieties of real-world materials. Due to its representability, we choose Disney's principled BSDF~\cite{DisneyPrincipledBSDF} which integrates five different BRDFs controlled by eleven parameters ({\it baseColor}, {\it subsurface}, {\it metallic}, {\it specular}, {\it specularTint}, {\it roughness}, {\it anisotropic}, {\it sheen}, {\it sheenTint}, {\it clearcoat}, {\it clearcoatGloss}). 
Since our target is isotropic materials without subsurface scattering, we neglect parameters such as {\it subsurface} and {\it anisotropic}. We also neglect {\it specularTint} that artistically colorizes the specularity and {\it clearcort} and {\it clearcoatGloss} that does not strongly affect the rendering results. While principled BSDF is effective, we found that there are some unrealistic combinations of parameters that we want to skip (\eg, metallic = 1 and roughness = 0, or metallic = 0.5). For avoiding those unrealistic parameters, we divide the entire parameter sets into three categories, (a) Diffuse, (b) Specular and (c) Metallic. We generate three datasets individually and evenly merge them when training the network. The value of each parameter is randomly selected within specific ranges for each parameter (see \Fref{fig:datageneration}-(a)). To realize spatially varying materials, we divide the object region in the rendered image into $P$ (\ie, 5000 for the training data) superpixels and use the same set of parameters at pixels within a superpixel (See~\Fref{fig:datageneration}-(b)).

For simulating complex light transport, we use {\it Cycles}~\cite{Cycles} renderer bundled in Blender~\cite{Blender}. The orthographic camera and the directional light are specified. For each rendering, we choose a set of an object, BSDF parameter maps (one for each parameter), and lighting configuration (\ie, Once roughly 1300 lights are uniformly distributed on the hemisphere, small random noises are added to each light). Once images were rendered, we create {\it CyclesPS} dataset by generating observation maps pixelwisely. For making the network robust to the test data of any number of images, observation maps are generated from a pixelwisely different number of images. Concretely, when generating an observation map, we pick a random subset of images whose number is whithin $50$ to $1300$ and whose corresponding elevation angle of the light direction is more than a random threshold value within $20$ to $90$ degrees.\footnote{The minimum number of images is 50 for avoiding too sparse observation map and we only picked the lights whose elevation angles were more than 20 degrees since it is practically less possible that the scene is illuminated from the side.} The training process takes 10 epochs for 150 image sets (\ie, 15 objects $\times$ 10 rotations for the rotational pseudo-invariance). Each image set contains around 50000 samples (\ie, number of pixels in the object mask).
\section{Experimental Results} \label{sec:experiment}
\begin{figure}[!t]
	\begin{center}
		\includegraphics[width=120mm]{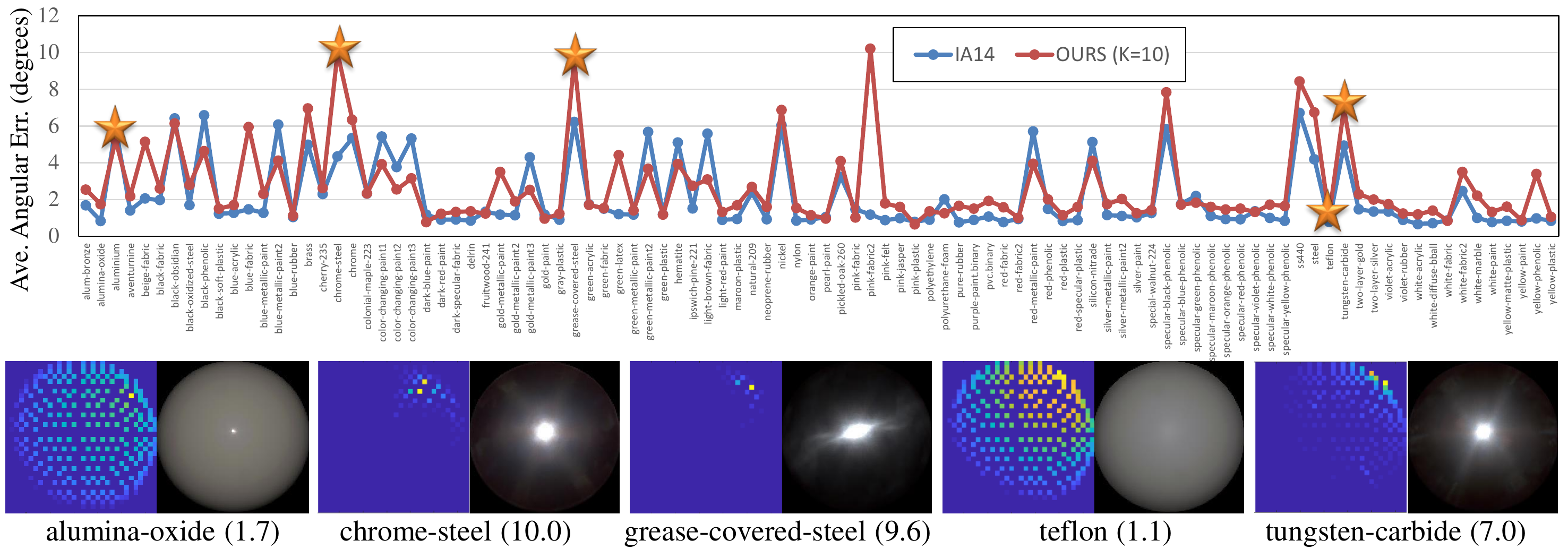}
	\end{center}
	\vspace{-5mm}
	\caption{Evaluation on the {\it MERLSphere} dataset. A sphere is rendered with 100 measured BRDF in MERL BRDF database~\cite{Matusik2003}. Our CNN-based method was compared against a model-based algorithm (IA14~\cite{ikehata2014a}) based on the  mean angular errors of predicted surface normals in degree. We also showed some examples of rendered images and observation maps for further analysis (See~\Sref{sec:experiment}.2).}
	\label{fig:merl}
\end{figure}
We evaluate our method on synthetic and real datasets. All experiments were performed on a machine with 3$\times$GeForce GTX 1080 Ti and 64GB RAM. For training and prediction, we use Keras library~\cite{chollet2015keras} with Tensorflow background and use default learning parameters. The training process took around 3 hours. 
\subsection{Datasets}
We evaluated our method on three datasets, two are synthetic and one is real. 

{\it MERLSphere} is a synthetic dataset where images are rendered with one hundred isotropic BRDFs in MERL database~\cite{Matusik2003} from diffuse to metallic. We generated 32-bit HDR images of a sphere ($256\times 256$) with a ground truth surface normal map and a foreground mask. There is no cast shadow and inter-reflection.

{\it CyclesPSTest} is a synthetic dataset of three objects, SPHERE, TURTLE and PAPERBOWL. TURTLE and PAPERBOWL are non-convex objects where the inter-reflection and cast shadow appear on rendered images. This dataset was generated in the same manner with the {\it CyclesPS} training dataset except that the number of superpixels in the parameter map was $100$ and the material condition was either Specular or Metallic (Note that objects and parameter maps in {\it CyclesPSTest} are NOT in {\it CyclesPS}). Each data contains 16-bit integer images with a resolution of $512\times 512$ under 17 or 305 known uniform lightings.

{\it DiLiGenT}~\cite{Shi2018} is a public benchmark dataset of 10 real objects of general reflectance. Each data provides 16-bit integer images with a resolution of $612\times 512$ from $96$ different known lighting directions. The ground truth surface normals for the orthographic projection and the single-view setup are also provided. 
\subsection{Evaluation on {\it MERLSphere} dataset}
We compared our method (with $K=10$ in~\Eref{eq:prediction}) against one of the state-of-the-art isotropic photometric stereo algorithms (IA14~\cite{ikehata2014b}\footnote{We used the authors' implementation of~\cite{ikehata2014b} with $N_1=2,N_2=4$ and turning on the retro-reflection handling. Attached shadows were removed by a simple thresholding. Note that our method takes into account all the input information unlike~\cite{ikehata2014b}.}) on the {\it MERLSphere} dataset. Without global illumination effects, we simply evaluate the ability of our network in representing wide varieties of materials compared to the sum-of-lobes BRDF~\cite{Chandraker2011a} introduced in IA14. The results are illustrated in \Fref{fig:merl}. We observed that our CNN-based algorithm performs comparably well, though not better than IA14, for most of materials, which indicates that 
Disney's principled BSDF~\cite{DisneyPrincipledBSDF} covers various real-world materials. We should note that as was commented in~\cite{DisneyPrincipledBSDF}, some of very shiny materials, particularly the metals (\eg, chrome-steel and tungsten-carbide), exhibited asymmetric highlights suggestive of lens flare or perhaps anisotropic surface scratches. Since our network was trained on purely isotropic materials, they inevitably degrade the performance.

\subsection{Evaluation on {\it CyclesPSTest} dataset}
\begin{table}[!t]
	\caption{Evaluation on the {\it CyclesPSTest} dataset. Here $m$ is the number of input images in each dataset and $\{S,M\}$ are types of material \ie, Specular (S) or Metallic (M) (See~\Fref{fig:datageneration} for details). For each cell, we show the average angular errors in degrees.}
	\begin{center}
		\includegraphics[width=120mm]{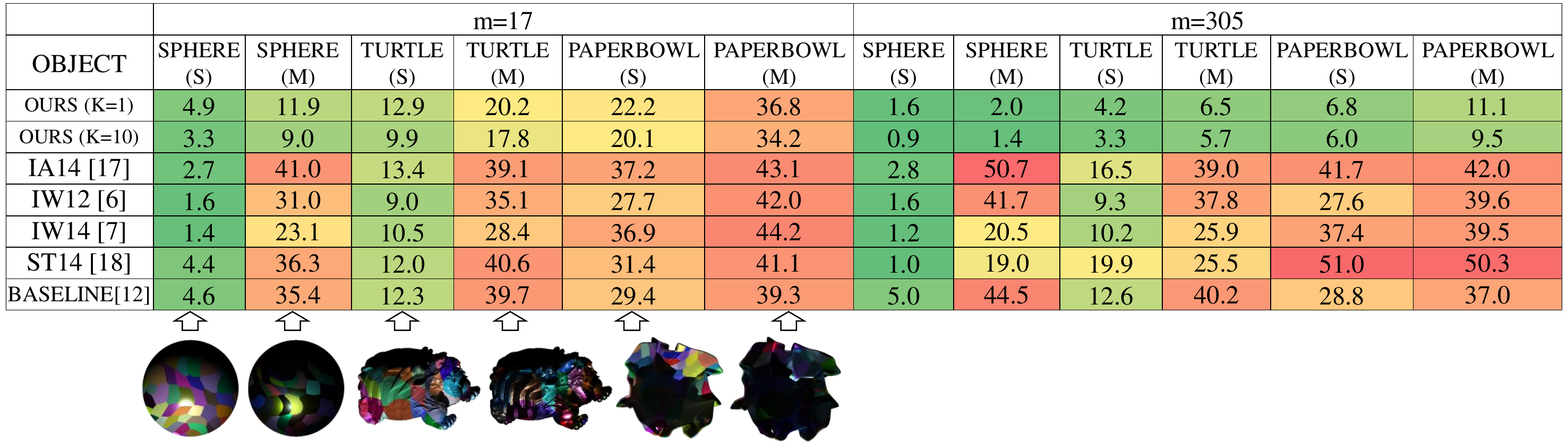}
	\end{center}
	\vspace{-5mm}	
	\label{table:prps}
\end{table}
\begin{table}[!t]
	\caption{Evaluation on the {\it DiLiGenT} dataset. We show the angular errors averaged within each object and over all the objects. (*) Our method discarded first 20 images in BEAR since they are corrupted (We explain about this issue in the supplementary).}
	\begin{center}
		\includegraphics[width=120mm]{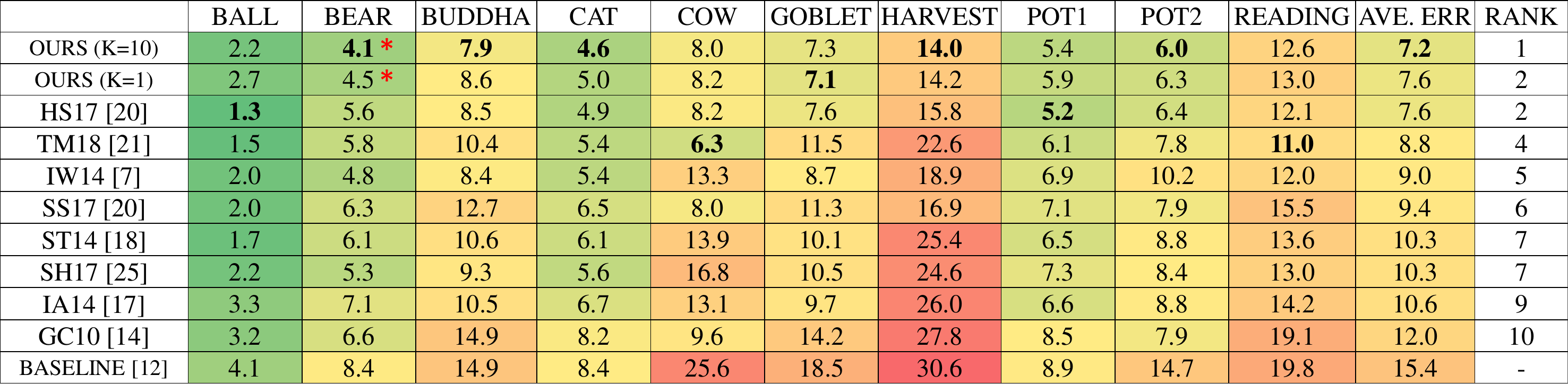}
	\end{center}
	\vspace{-5mm}	
	\label{table:diligent}
\end{table}

\begin{figure}[t]
	\begin{center}
		\includegraphics[width=120mm]{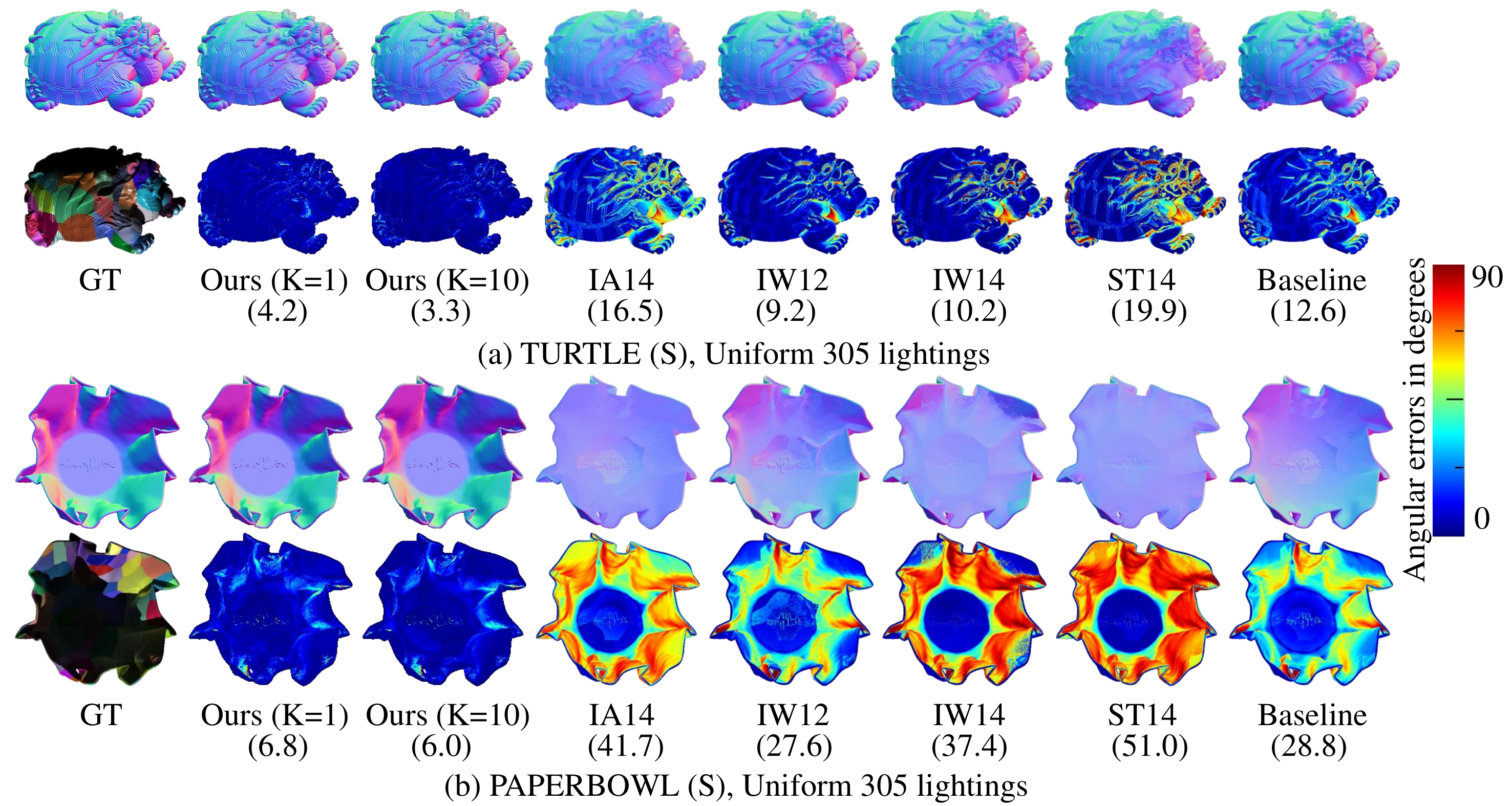}
	\end{center}
	\vspace{-5mm}
	\caption{Recovered surface normals and error maps for (a) TURTLE and (b) PAPERBOWL of Specular material. Images were rendered under uniform 305 lightings.}
	\label{fig:prps}
\end{figure}

To evaluate the ability of our method in recovering non-convex surfaces, we tested our method on {\it CyclesPSTest}. Our method was compared against two robust algorithms IW12~\cite{ikehata2012} and IW14~\cite{ikehata2014a}\footnote{We used authors' implementation and set parameters of~\cite{ikehata2012} as $\lambda=0,\sigma=1.0^{-6}$ and parameters of~\cite{ikehata2014a} as $\lambda=0,p=3,\sigma_a=1.0$.}, two model-based algorithms ST14~\cite{Shi2014}\footnote{We used our implementation of~\cite{Shi2014} and set $T_{low}=0.25$.} and IA14~\cite{ikehata2014b} and BASELINE~\cite{Woodham1980}. When running algorithms except for ours, we discarded samples whose intensity values were less than $655$ in a 16-bit integer image for the shadow removal. In this experiment, we also studied the effects of number of images and  rotational merging in the prediction.\footnote{We still augument data by rotations in the training step.} Concretely, we tested our method on 17 or 305 images with $K=1$ and $K=10$ in~\Eref{eq:prediction}. We show the results in~\Tref{table:prps} and~\Fref{fig:prps}. We observed that all the algorithms worked well on the convex specular SPHERE dataset. However, when surfaces were non-convex, all the algorithms except ours failed in the estimation due to strong cast shadow and inter-reflections. It is interesting to see that even the robust algorithms (IA12~\cite{ikehata2012} and IA14~\cite{ikehata2014a}) could not deal with the global effects as outliers. We also observed that the rotational averaging based on the rotational pseudo-invariance definitely improved the accuracy, though not very much. 
\subsection{Evaluation on {\it DiLiGenT} dataset} 
\begin{figure}[t]
	\begin{center}
		\includegraphics[width=120mm]{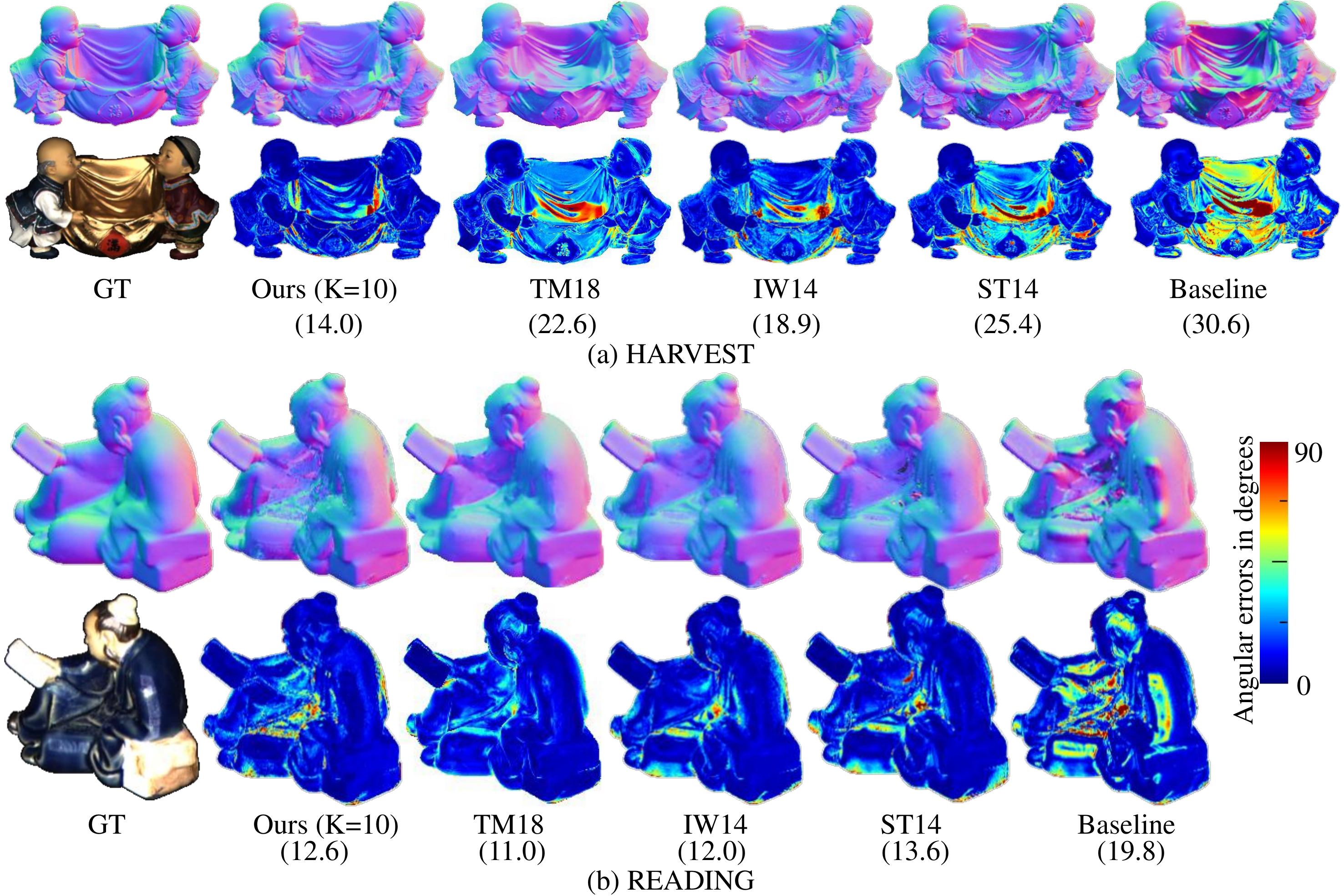}
	\end{center}
	\vspace{-5mm}
	\caption{Recovered surface normals and error maps for (a) HARVEST and (b) READING in the {\it DiLiGenT} dataset.}
	\label{fig:diligent}
\end{figure}

Finally, we present a side-by-side comparison on the {\it DiLiGenT} dataset~\cite{Shi2018}. We collected existing benchmark results for the calibrated photometric stereo algorithms~\cite{Woodham1980,Alldrin2008,Goldman2010,Higo2010,Wu2010,ikehata2012,Shi2012b,ikehata2014a,ikehata2014b,Shi2014,Queau2017,Santo2017,Hui2017,taniai2018}. Note that we compared the mean angular errors of~\cite{Woodham1980,Alldrin2008,Goldman2010,Higo2010,Wu2010,Shi2012b,ikehata2014b,Shi2014} reported in~\cite{Shi2018}, ones reported in their own works~\cite{Santo2017,Hui2017,taniai2018} and ones from our experiment using authors' implementation~\cite{ikehata2012,ikehata2014a,Queau2017}.\footnote{As for~\cite{Queau2017}, we used the default setting of their package except that we gave the camera intrinsics provided by~\cite{Shi2018} and changed the noise variance to zero.} The results are illustrated in~\Tref{table:diligent}. Due to the space limit, we only show the top-10 algorithms\footnote{Please find the full comparison in our supplementary.} w.r.t the overall mean angular, and BASELINE~\cite{Woodham1980}. We observed that our method achieved the smallest errors averaged over 10 objects, best scores for 6 of 10 objects. It is valuable to note that other top-ranked algorithms~\cite{Hui2017,taniai2018} are time-consuming since HS17~\cite{Hui2017} requires the dictionary learning for every different light configuration and TM18~\cite{taniai2018} needs the unsupervised training for every estimation while our inference time is less than five seconds (when $K=1$) for each dataset on CPU. Taking a close look at each object, \Fref{fig:diligent} provides some important insights. HARVEST is the most non-convex scene in {\it DiLiGenT} and other state-of-the art algorithms (TM18~\cite{taniai2018}, IW14{\cite{ikehata2014a}}, ST14~\cite{Shi2014}) failed in the estimation of normals inside the ``bag" due to strong shadow and inter-reflections. Our CNN-based method estimated much more reasonable surface normals there thanks to the network trained based on the carefully created {\it CyclesPS} dataset. On the other hand, our method did not work best (though not bad) for READING which is another non-convex scene. Our analysis indicated that this is because of the inter-reflection of {\it  high-intensity narrow specularities} that were rarely observed in our training dataset (Narrow specularities appear only when {\it roughness} in the principled BSDF is near zero).
\section{Conclusion}
In this paper, we have presented a CNN-based photometric stereo method which works for various kind of isotropic scenes with global illumination effects. By projecting photometric images and lighting information onto the observation map, unstructured information is naturally fed into the CNN. Our detailed experimental results have shown the state-of-the-art performance of our method for both synthetic and real data especially when the surface is non-convex. To make better training set for handling narrow inter-reflections is our future direction.
\clearpage

\bibliographystyle{splncs}
\bibliography{egbib}

\begin{thebibliography}{10}

\bibitem{Kendall2017}
Kendall, A., Martirosyan, H., Dasgupta, S., Henry, P., Kennedy, R., Bachrach,
  A., Bry, A.:
\newblock End-to-end learning of geometry and context for deep stereo
  regression.
\newblock {Proc. ICCV} (2017)

\bibitem{Vijayanarasimhan2017}
Vijayanarasimhan, S., Ricco, S., Schmid, C., Sukthankar, R., Fragkiadaki, K.:
\newblock Sfm-net: Learning of structure and motion from video.
\newblock arXiv preprint arXiv:1704.07804 (2017)

\bibitem{Kar2017}
Kar, A., H\"{a}ne, C., Malik, J.:
\newblock Learning multi-view stereo machine.
\newblock {Proc. NIPS} (2017)

\bibitem{Kim2017}
Kim, K., Gu, J., Tyree, S., Molchanov, P., Niessner, M., Kautz, J.:
\newblock A lightweight approach for on-the-fly reflectance estimation.
\newblock {Proc. ICCV} (2017)

\bibitem{Wu2010}
Wu, L., Ganesh, A., Shi, B., Matsushita, Y., Wang, Y., Ma, Y.:
\newblock Robust photometric stereo via low-rank matrix completion and
  recovery.
\newblock In: {Proc. ACCV}. (2010)

\bibitem{ikehata2012}
Ikehata, S., Wipf, D., Matsushita, Y., Aizawa, K.:
\newblock Robust photometric stereo using sparse regression.
\newblock In: {Proc. CVPR}. (2012)

\bibitem{ikehata2014a}
Ikehata, S., Wipf, D., Matsushita, Y., Aizawa, K.:
\newblock Photometric stereo using sparse bayesian regression for general
  diffuse surfaces.
\newblock {IEEE Trans. Pattern Anal. Mach. Intell.} \textbf{36}(9) (2014)
  1816--1831

\bibitem{Queau2017}
Quéau, Y., Wu, T., Lauze, F., Durou, J.D., Cremers, D.:
\newblock A non-convex variational approach to photometric stereo under
  inaccurate lighting.
\newblock In: {Proc. CVPR}. (2017)

\bibitem{Cycles}
Cycles.
\newblock https://www.cycles-renderer.org/

\bibitem{DisneyPrincipledBSDF}
Burley, B.:
\newblock Physically-based shading at disney, part of practical physically
  based shading in film and game production.
\newblock SIGGRAPH 2012 Course Notes (2012)

\bibitem{Shi2018}
Shi, B., Mo, Z., Wu, Z., D.Duan, Yeung, S.K., Tan, P.:
\newblock A benchmark dataset and evaluation for non-lambertian and
  uncalibrated photometric stereo.
\newblock {IEEE Trans. Pattern Anal. Mach. Intell.} (2018)  (to appear)

\bibitem{Woodham1980}
Woodham, P.:
\newblock Photometric method for determining surface orientation from multiple
  images.
\newblock Opt. Engg \textbf{19}(1) (1980)  139--144

\bibitem{Alldrin2008}
Alldrin, N., Zickler, T., Kriegman, D.:
\newblock Photometric stereo with non-parametric and spatially-varying
  reflectance.
\newblock In: {Proc. CVPR}. (2008)

\bibitem{Goldman2010}
Goldman, D.B., Curless, B., Hertzmann, A., Seitz, S.M.:
\newblock Shape and spatially-varying brdfs from photometric stereo.
\newblock {IEEE Trans. Pattern Anal. Mach. Intell.} \textbf{32}(6) (2010)
  1060--1071

\bibitem{Higo2010}
Higo, T., Matsushita, Y., Ikeuchi, K.:
\newblock Consensus photometric stereo.
\newblock In: {Proc. CVPR}. (2010)

\bibitem{Shi2012b}
Shi, B., Tan, P., Matsushita, Y., Ikeuchi, K.:
\newblock Elevation angle from reflectance monotonicity.
\newblock In: {Proc. ECCV}. (2012)

\bibitem{ikehata2014b}
Ikehata, S., Aizawa, K.:
\newblock Photometric stereo using constrained bivariate regression for general
  isotropic surfaces.
\newblock In: {Proc. CVPR}. (2014)

\bibitem{Shi2014}
Shi, B., Tan, P., Matsushita, Y., Ikeuchi, K.:
\newblock Bi-polynomial modeling of low-frequency reflectances.
\newblock {IEEE Trans. Pattern Anal. Mach. Intell.} \textbf{36}(6) (2014)
  1078--1091

\bibitem{Santo2017}
Santo, H., Samejima, M., Sugano, Y., Shi, B., Matsushita, Y.:
\newblock Deep photometric stereo network.
\newblock In: International Workshop on Physics Based Vision meets Deep
  Learning (PBDL) in Conjunction with IEEE International Conference on Computer
  Vision (ICCV). (2017)

\bibitem{Hui2017}
Hui, Z., Sankaranarayanan, A.C.:
\newblock Shape and spatially-varying reflectance estimation from virtual
  exemplars.
\newblock {IEEE Trans. Pattern Anal. Mach. Intell.} \textbf{39}(10) (2017)
  2060--2073

\bibitem{taniai2018}
Taniai, T., Maehara, T.:
\newblock {Neural Inverse Rendering for General Reflectance Photometric
  Stereo}.
\newblock In: {Proc. ICML}. (2018)

\bibitem{Goldman2005}
Goldman, D., Curless, B., Hertzmann, A., Seitz, S.:
\newblock Shape and spatially-varying brdfs from photometric stereo.
\newblock In: {Proc. ICCV}. (October 2005)

\bibitem{Ward1992}
Ward, G.:
\newblock Measuring and modeling anisotropic reflection.
\newblock Computer Graphics \textbf{26}(2) (1992)  265--272

\bibitem{Chandraker2011a}
Chandraker, M., Ramamoorthi, R.:
\newblock What an image reveals about material reflectance.
\newblock In: {Proc. ICCV}. (2011)

\bibitem{Shen2017}
Shen, H.L., Han, T.Q., Li, C.:
\newblock Efficient photometric stereo using kernel regression.
\newblock IEEE Transactions on Image Processing \textbf{26}(1) (2017)  439--451

\bibitem{Silver1980}
Silver, W.M.:
\newblock Determining shape and reflectance using multiple images.
\newblock Master's thesis, MIT (1980)

\bibitem{Hertzmann2005}
Hertzmann, A., Seitz, S.:
\newblock Example-based photometric stereo: shape reconstruction with general,
  varying brdfs.
\newblock {IEEE Trans. Pattern Anal. Mach. Intell.} \textbf{27}(8) (2005)
  1254--1264

\bibitem{Huan2017}
G.~Huang, Z.~Liu, L.M.K.W.:
\newblock Densely connected convolutional networks.
\newblock In: {Proc. CVPR}. (2017)

\bibitem{Matusik2003}
Matusik, W., Pfister, H., Brand, M., McMillan, L.:
\newblock A data-driven reflectance model.
\newblock {ACM Trans. on Graph.} \textbf{22}(3) (2003)  759--769

\bibitem{Alldrin2007b}
Alldrin, N., Kriegman, D.:
\newblock Toward reconstructing surfaces with arbitrary isotropic reflectance:
  A stratified photometric stereo approach.
\newblock In: {Proc. ICCV}. (2007)

\bibitem{Stark2005}
Stark, M., Arvo, J., Smits, B.:
\newblock Barycentric parameterizations for isotropic brdfs.
\newblock IEEE Trans. on Visualization and Computer Graphics \textbf{11}(2)
  (2011)  126--138

\bibitem{Montes2012}
Montes, R., Urena, C.:
\newblock An overview of brdf models.
\newblock Technical report, LSI-2012-001 en Digibug Coleccion: TIC167 -
  Articulos (2012)

\bibitem{Simard2003}
Simard, P.Y., Steinkraus, D., Platt, J.C.:
\newblock Best practices for convolutional neural networks applied to visual
  document analysis.
\newblock In Proc. ICDAR (2003)

\bibitem{Schmidt2012}
Schmidt, U., Roth, S.:
\newblock Learning rotation-aware features: From invariant priors to
  equivariant descriptors.
\newblock {Proc. CVPR} (2012)

\bibitem{Adam}
Kingma, D., Ba, J.:
\newblock Adam: A method for stochastic optimization.
\newblock In proc. ICLR (2014)

\bibitem{shapenet2015}
Chang, A.X., Funkhouser, T., Guibas, L., Hanrahan, P., Huang, Q., Li, Z.,
  Savarese, S., Savva, M., Song, S., Su, H., Xiao, J., Yi, L., Yu, F.:
\newblock {ShapeNet: An Information-Rich 3D Model Repository}.
\newblock Technical Report arXiv:1512.03012 [cs.GR], Stanford University ---
  Princeton University --- Toyota Technological Institute at Chicago (2015)

\bibitem{Blender}
Blender.
\newblock https://www.cycles-renderer.org/

\bibitem{chollet2015keras}
Chollet, F.,  et~al.:
\newblock Keras.
\newblock \url{https://github.com/keras-team/keras} (2015)

\end{thebibliography}
\end{document}